\documentclass{llncs}
\usepackage{makeidx,nth}  
\usepackage{url}
\usepackage{graphicx,tabularx,diagbox}
\usepackage{booktabs,colortbl}
\usepackage{subfig}
\usepackage[group-separator={,}]{siunitx}
\usepackage[table]{xcolor}
\usepackage{pgfplots}
\usepackage[]{nth}
\usepackage{listings}

\DeclareFixedFont{\ttb}{T1}{txtt}{bx}{n}{8} 
\DeclareFixedFont{\ttm}{T1}{txtt}{m}{n}{8}  

\usepackage{color}
\definecolor{deepblue}{rgb}{0,0,0.5}
\definecolor{deepred}{rgb}{0.6,0,0}
\definecolor{deepgreen}{rgb}{0,0.5,0}

\newcommand\pythonstyle{\lstset{
language=Python,
basicstyle=\ttm,
otherkeywords={self},             
keywordstyle=\ttb\color{deepblue},
emph={MyClass,__init__},          
emphstyle=\ttb\color{deepred},    
stringstyle=\color{deepgreen},
frame=tb,                         
showstringspaces=false            %
}}

\lstnewenvironment{python}[1][]
{
\pythonstyle
\lstset{#1}
}
{}

\usepgfplotslibrary{groupplots}
\pgfplotsset{compat=newest}
\newcolumntype{a}{>{\columncolor[gray]{0.9}}c}
\newcolumntype{b}{>{\columncolor[gray]{0.8}}c}

\newcommand\pythoninline[1]{{\pythonstyle\lstinline!#1!}}

\begin{document}
\title{DELIMIT PyTorch - An extension for Deep Learning in Diffusion Imaging}
\titlerunning{Deep Learning in Diffusion Imaging}

\author{Simon Koppers \and Dorit Merhof}
\institute{RWTH Aachen University, Germany\\
contact email: koppers@lfb.rwth-aachen.de}
\maketitle              

\begin{abstract}
DELIMIT is a framework extension for deep learning in diffusion imaging, which extends the basic framework PyTorch towards spherical signals. 
Based on several novel layers, deep learning can be applied to spherical diffusion imaging data in a very convenient way.
First, two spherical harmonic interpolation layers are added to the extension, which allow to transform the signal from spherical surface space into the spherical harmonic space, and vice versa.
In addition, a local spherical convolution layer is introduced that adds the possibility to include gradient neighborhood information within the network.
Furthermore, these extensions can also be utilized for the preprocessing of diffusion signals. 
\keywords{Diffusion MRI, Magnetic Resonance Imaging, Machine Learning, Deep Learning, Spherical Harmonic, Spherical Convolution}
\end{abstract}
\section{Introduction}
Diffusion imaging (DI) rapidly developed into one of the most important non-invasive tools for clinical brain research, due to its ability to reconstruct neural pathways in the human brain.
However, long acquisition times, based on the high amount of acquired gradient directions, result in a rare usage of DI in clinical practice. 
To overcome this problem, recent methods demonstrated the strength of machine learning and in particular deep learning (DL), which is able to describe and reconstruct the tissue's underlying complex microstructure very accurately even if only a limited number of gradient directions are available~\cite{Golkov2015,Koppers2016}. 
Thus, scanning time can be greatly reduced.

Despite these promising results, the application of DL in the field of DI is constrained, due to a non-uniform signal representation, resulting from diverging gradient directions and b-values during acquisition across subjects.
Even if the same gradient directions remain identical within a study, scanner artifacts, subject motion or eddy currents will lead to distorted gradient directions~\cite{Andersson2003,Andersson2016}.
To address this issue, a uniform and unique signal representation is necessary to apply DL in DI.
Therefore, all signals must be transformed into a generic representation, e.g. spherical harmonics (SH)~\cite{Descoteaux2007}.

This is further supported by the fact that SH coefficients normally average around zero, which is preferred for DL applications~\cite{Ioffe2015}.
On the other hand, this representation is very susceptible to noise, as high orders commonly have very small coefficients.
Therefore, fine structures and sharp contrasts within the signal cannot be learned, resulting in a smoothing effect on the signal.

A simple way to address this issue could be to transform the signal back into the spherical surface space (which would be the q-space in case of DI), since all gradient directions can be assumed to be weighted equally.
Furthermore, transforming the signal from SH into spherical surface space also opens the application of different loss functions, including relative approaches, which cannot be utilized if the average predicted value is close to zero.


Another issue of spherical signals in DL is that spatial convolutions assume all gradient directions to be independent, while gradient neighborhood information is discarded.
Furthermore, spatial convolutions merge multi-shell signals, which consist of multiple b-values, into a single signal.
To make use of this additional information, a novel layer for DL in DI is introduced: the local spherical convolution (LSC). 
This convolution utilizes cyclic kernels, which are applied to the surface of a spherical diffusion signal.
Theses kernels are also extended to multi-shell signals to improve multi-shell predictions.

To allow these extensions to be utilized in the field of deep learning, we implemented these layers efficiently on the GPU.
The code for this paper is publicly available upon request.
%
%
%
\section{DELImIt Extension}
%
DELIMIT is a framework extension for DL in DI, which extends the basic framework PyTorch towards spherical signals. 
These extensions can also be utilized for the preprocessing of diffusion signals. 

Its core elements are signal transformation layer, transforming a spherical signal from the surface space into the SH space, and vice versa (see \pythoninline{class Signal2SH} and \pythoninline{class SH2Signal}).
Furthermore, a local spherical convolution layer is added, which includes gradient neighborhood information within training and application of neural networks.
\subsection{The basic concept of PyTorch}
Since DELIMIT extends the PyTorch framework for an application of DL in DI, its basic element is a tensor.
This tensor can be seen as a multidimensional matrix, while it allows to perform different operations efficiently on a GPU.
Within this toolbox every tensor has five dimensions, which are defined by $size(tensor) = \mathrm{subjects} \times \mathrm{\#shells*\#gradient directions} \times \mathrm{height} \times \mathrm{width} \times \mathrm{depth}$.
Further examples on how to generate and how to use a tensor are given in the individual class section.
\subsection{Spherical Harmonics}
The most important extension within DELIMIT are \pythoninline{class Signal2SH} and \pythoninline{class SH2Signal}, which convert the diffusion signal into harmonic space, and vice versa.
\subsubsection{Signal2SH}
This class converts a signal from the spherical surface space into SH space.
It requires the signal to be normalized based on its non-diffusion weighted $b=0\,\frac{\mathrm{s}}{\mathrm{mm}^2}$ measurement.
\paragraph{Parameters:}
\begin{itemize}
	\item sh\_order $($int$)$: SH order
	\item gradients (float, $N\times$3 matrix): acquired $N$ gradient directions
	\item lb\_lambda $($float, \textit{optional}$)$: Laplace-Beltrami regularization parameter
\end{itemize}
\paragraph{Example:\newline}
\begin{python}
data, affine = load_nifti('data.nii.gz')
bvals, bvecs = read_bvals_bvecs('bvals', 'bvecs')
gradients = bvecs[bvals > 0, :]
b0_data = np.mean(data[:, :, :, bvals==0], axis=3)
data = data[:, :, :, bvals > 0] / np.expand_dims(b0_data, 3)

s2sh = Signal2SH(sh_order=4, gradients=gradients, lb_lambda=0.06)
signal = Variable(torch.from_numpy(data))
signal = signal.contiguous().permute(3, 0, 1, 2).cuda()
data_sh = s2sh(signal.unsqueeze(0)).squeeze()
\end{python}
\subsubsection{SH2Signal}
This class converts a signal from SH space back into the spherical surface space.
It requires the signal to be normalized based on its non-diffusion weighted $b=0\,\frac{\mathrm{s}}{\mathrm{mm}^2}$ measurement.
\paragraph{Parameters:}
\begin{itemize}
	\item sh\_order (int): SH order
	\item gradients (float, $N\times$3 matrix): resampled $N$ gradient directions
\end{itemize}
\paragraph{Example:\newline}
\begin{python}
sh2s = SH2Signal(sh_order=4, gradients=bvecs[bvals > 0, :])
data_reconstruced = sh2s(data_sh.unsqueeze(0)).squeeze()
data_reconstruced = data_reconstruced.permute(1, 2, 3, 0).cpu().numpy()
\end{python}

\subsubsection{Efficiency of Signal2SH and SH2Signal}
\begin{figure}[ht]
	\subfloat[Spherical surface space into SH space]{%
		\includegraphics[width=0.48\textwidth]{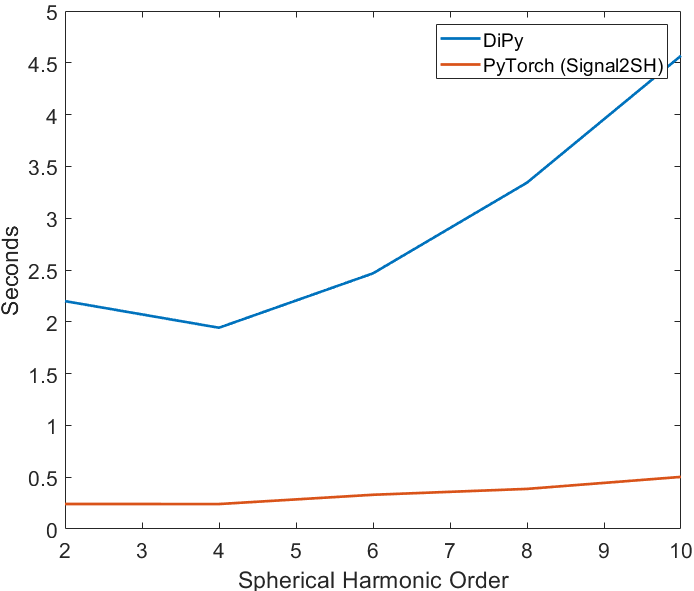}
	}
	\hfill
	\subfloat[SH space into spherical surface space]{%
		\includegraphics[width=0.48\textwidth]{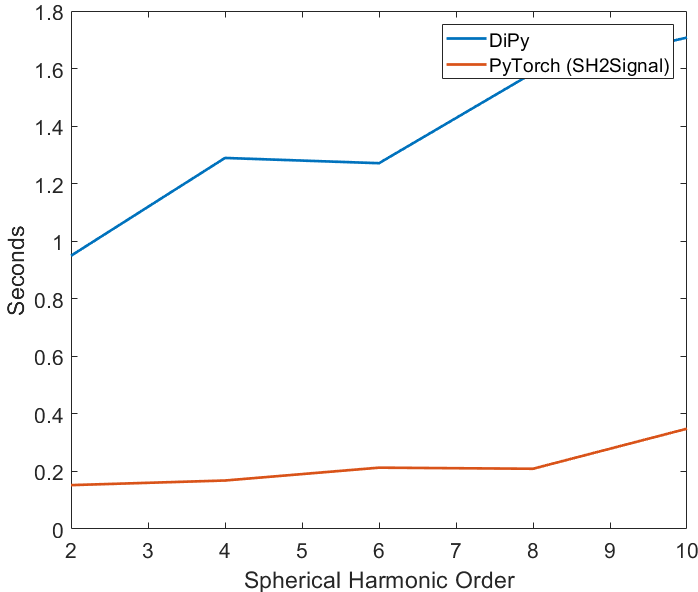}
	}
	\caption{Evaluation of transforming a full brain scan ($\approx$~450,000 voxels.) from SH space into the corresponding spherical surface space, and vice versa. It is evaluated for different SH orders.}
	\label{fig:SHEva}
\end{figure}

In order to apply SH within a neural network, a requirement is that it does not slow down the training of a neural network nor its application. 
For this purpose the \pythoninline{class Signal2SH} and \pythoninline{class SH2Signal} computational performance is evaluated for different SH orders on a dataset with $\approx$~450,000 voxels (see Fig.~\ref{fig:SHEva}). 
As a comparison, the same task was performed with the \pythoninline{sf_to_sh} and its corresponding \pythoninline{sh_to_sf} function, which are efficiently implemented as a vector operation in dipy~\cite{Dipy}.
It can be seen that the PyTorch transformation speeds-up the transformation by a factor of $\approx$~9, while both are mathematically identical (apart from numerical differences due to \textit{float} and \textit{double} precision.
\subsection{Local Spherical Convolution}
\begin{figure}
	\centering
		\includegraphics[width=0.5\textwidth]{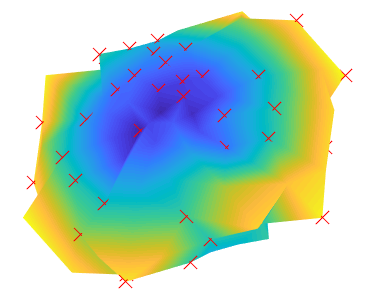}
	\caption{Acquired gradient directions (marked with red crosses), which span a spherical signal.}
	\label{fig:sampled_signal}
\end{figure}

While spatial neighborhood information, defined by neighboring voxels, is commonly utilized in DL based on convolutional layers, local spherical neighborhood information gets discarded during spatial convolution.
To address this issue, the LSC layer was developed for spherical signals that occur in DI as long as spherical acquisition schemes are utilized. 
In such a case, each diffusion signal spans a sphere based on its acquired gradient directions (see Fig.~\ref{fig:sampled_signal}).
However, this type of convolution is mathematically not defined.
The only defined convolution in the spherical region is a convolution of two complete spherical signals, which is not the desired way of convolution in this case.

\begin{figure}[ht]
	\centering
		\includegraphics[width=0.75\textwidth]{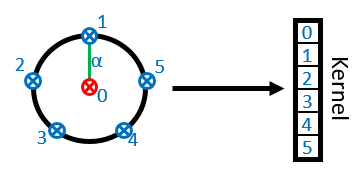}
	\caption{A circular-shaped kernel with an angular distance $\alpha$ and $n=5$.}
	\label{fig:kernel}
\end{figure}

In the context of a local spherical convolution, a circular-shaped kernel is convoluted over the signal's surface, spanned by adjacent gradient directions.
These kernels are defined by an angular distance $\alpha$, which defines the angle between the origin and every other point within the kernel.
Furthermore, the number of additional kernel points $n$ on a circle around the origin needs to be defined.
An exemplary kernel with $n=5$ can be seen in Fig.~\ref{fig:kernel}.
After definition, it is stored in a 1D vector with $n+1$ entries.

Due to individual gradient directions differences, each gradient's neighborhood is resampled based on a SH interpolation. 
This is also necessary, since not every gradient direction has the same amount of neighboring gradient directions.
Therefore, $n$ new gradient directions need to be resampled for every initially acquired gradient direction.
This ensures that neighboring gradient directions have the same angular distance in comparison to the kernel's origin. 

\begin{figure}[ht]
	\centering
		\includegraphics[width=0.95\textwidth]{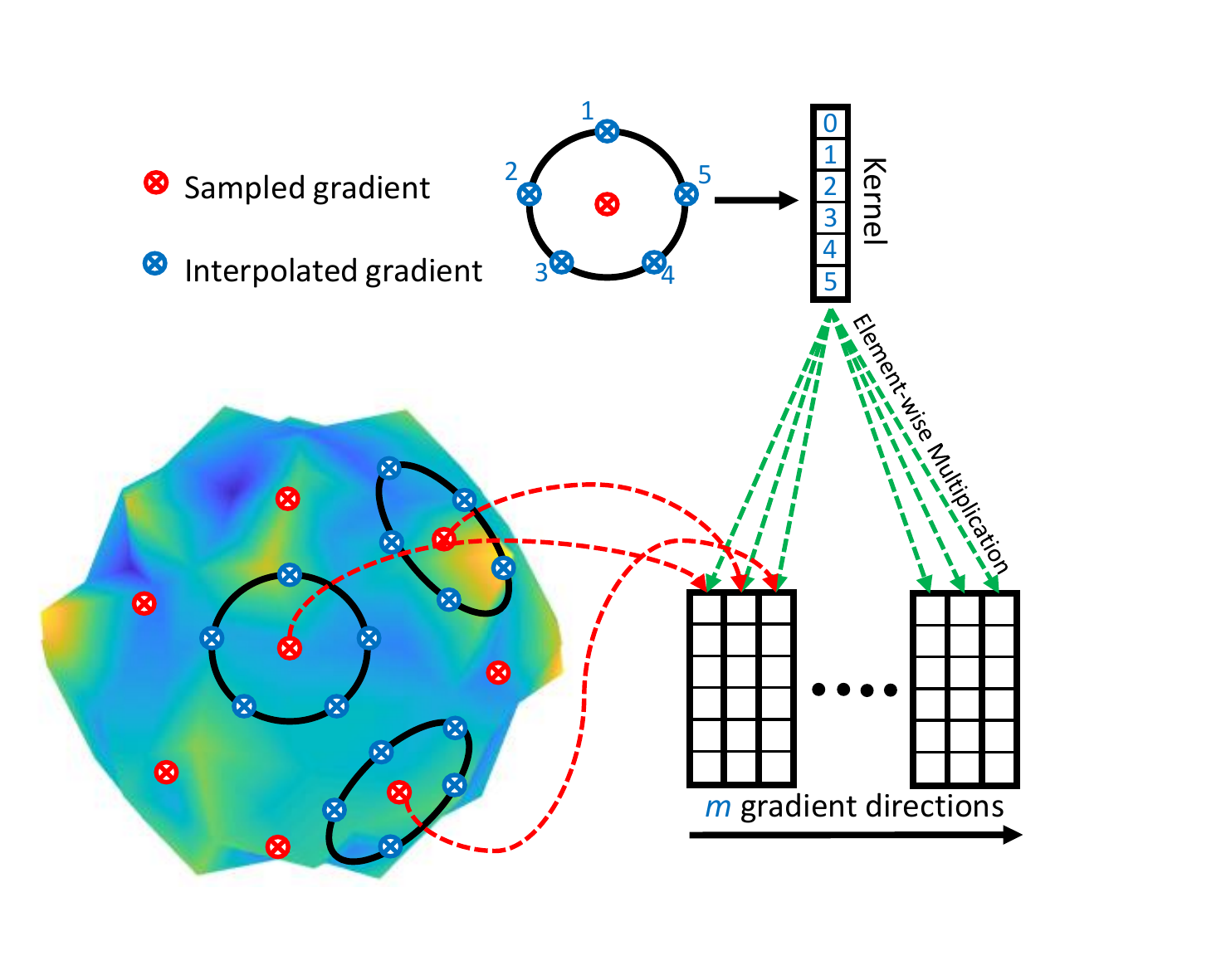}
	\caption{Workflow of a Spherical Convolution.}
	\label{fig:workflow}
\end{figure}

\begin{figure}[ht]
	\centering
	\subfloat[Raw Inputsignal.]{%
		\includegraphics[width=0.3\textwidth]{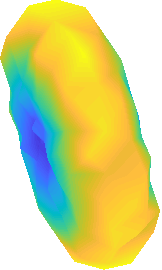}
	}
	\hspace{3cm}
	\subfloat[Signal after local spherical convolution.]{%
		\includegraphics[width=0.3\textwidth]{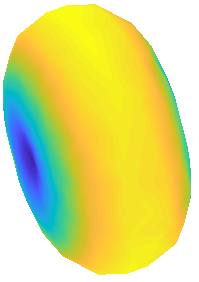}
	}
	\caption{Visualization of a diffusion signal convoluted with a moving average filter of size $n=5$ and $\alpha = \frac{\pi}{5}$.}
	\label{fig:signal_sconv}
\end{figure}

After resampling, every origin and its resampled neighborhood is stored in a 1D vector, resulting in a $m\times (n+1)$ matrix, where $m$ is the number of original gradient directions and $n+1$ is the size of the kernel. 
In the end, the local spherical convolution is performed by point-wise multiplying the 1D kernel vector to every 1D vector, resulting in $m$ scalar values, which can be seen as a local spherical convolution of the input signal.
The full pipeline is also shown in Fig.~\ref{fig:workflow}.
In addition, an example of a convoluted signal is shown in Fig.~\ref{fig:signal_sconv}.
Here, a moving average filter with $n=5$ and an angular distance of $\alpha=\frac{\pi}{5}$ is utilized.
It can be seen that a moving average filter leads to a expected blurred version of the initial diffusion signal.
\newpage
\paragraph{Parameters:}
\begin{itemize}
	\item shells\_in (int): number of input shells
	\item shells\_out (int): number of output shells
	\item sh\_order\_in (int): SH order of the input signal
	\item sh\_order\_out (int): SH order of the output signal
	\item sampled\_gradients (float, $N\times$3 matrix): utilized $N$ gradient directions
	\item kernel\_sizes (list): list of kernel size
	\item lb\_lambda (float, \textit{optional}): number of output shells
	\item angular\_distance (float, \textit{optional}): number of output shells
\end{itemize}

\paragraph{Example:\newline}
\begin{python}
# Data Loader
data_tmp, affine = load_nifti('data_b1000.nii.gz')
bvals, bvecs = read_bvals_bvecs('bvals', 'bvecs')
sh_order = 4
bvecs = bvecs[bvals == 1000, :]
bvals = bvals[bvals == 1000]

# Signal Transformation Layer
s2sh = Signal2SH(sh_order=sh_order, gradients=bvecs, lb_lambda=0.006)
sh2s = SH2Signal(sh_order=sh_order, gradients=bvecs)

signal = torch.from_numpy(data_tmp).contiguous().permute(3, 0, 1, 2)
if torch.cuda.is_available():
    signal = signal.cuda()
		
# Initialization of Local Spherical Convolution
lsc = LocalSphericalConvolution(
shells_in=1, 
shells_out=1,
sampled_gradients=bvecs, 
sh_order_in=sh_order, 
sh_order_out=sh_order, 
kernel_sizes=[5], 
angular_distance=np.pi/5, 
lb_lambda=0.006)

# Definition of kernel; Bias is set to zero
moving_average_kernel = torch.Tensor([[[[1, 1, 1, 1, 1, 1]]]])
moving_average_kernel /= torch.sum(moving_average_kernel)
lsc.sconv.weight  = torch.nn.Parameter(moving_average_kernel)
lsc.sconv.bias  = torch.nn.Parameter(torch.Tensor([0]))

# Apply Convolution
sh = s2sh(signal.unsqueeze(0))
sh_conv = lsc(sh)
signal_conv = sh2s(sh_conv)
\end{python}
Additionally, it should be mentioned that in the context of DL these operations are falsely called ``convolution", whereas actually cross correlations are meant.
During training of a neural network, however, both operations lead to the same result, while the implementation of a cross correlation is significantly faster.

\subsubsection{Extension to multiple b-values}
The LSC can also be applied to multiple b-values.
In this case, an additional channel dimension is assigned to the kernel, which determines the weight of each input channel (or in case of DI every b-value).
Furthermore, multiple b-values can be predicted, by employing multiple kernels to the input signal, whereas each kernel leads to its own b-value.

\section{Discussion}
Current DL frameworks offer users a convenient and fast way to apply machine learning and DL to any kind of data. 
However, these general toolboxes do not extend to specialized application, e.g. DI, creating several problems:
\begin{enumerate}
	\item No spherical signal transformation layers to switch between the spherical surface space and the SH space.
	\item Discarding of spherical information during spatial convolution
\end{enumerate}
We addressed these issues by developing DELIMIT, an open-source extension for PyTorch, introducing three novel layers for DL in DI. 
As shown in Fig.~\ref{fig:SHEva} it can be seen that the implemented transformation layers (\pythoninline{class Signal2SH} and \pythoninline{class SH2Signal}) are very fast and efficient, especially as the SH order increases of if more voxels are transformed.
Furthermore, a local spherical convolution layer is introduced, which is able to include additional gradient neighboring information during training and application of neural networks.
It is also able to process multi-shell signals, which are acquired with different b-values.

In general, it should be noted that all proposed layers require a powerful GPU with a large amount of GPU RAM, if full brain datasets are processed.
As an example: A full Human Connectom Project dataset (288 gradient directions with isotropic voxel resolution of 1.25mm$^3$, at least 2~GB RAM is required for the dataset without any layers being loaded.
\section{Conclusion}
In summary, this toolbox is very convenient to use and could significantly contribute to the DI community in the field of DL.
Furthermore, it is possible to use this toolbox for processing of diffusion signals without utilizing DL. 
In such a case, a significant acceleration of existing methods could be be achieved.

As a next step, we plan to add further existing DI methods to DELIMIT.
\bibliographystyle{splncs03}
\bibliography{koppers}
\end{document}